\documentclass{llncs}
\usepackage[english]{babel}
\usepackage[utf8x]{inputenc}
\usepackage[T1]{fontenc}
\usepackage{color,soul}

\usepackage{amsmath}
\usepackage{amssymb}
\usepackage{graphicx}
\usepackage[colorinlistoftodos]{todonotes}
\usepackage[colorlinks=true, allcolors=blue]{hyperref}
\usepackage{algorithm}
\usepackage{algorithmic}

\title{Brenier approach for optimal transportation between a quasi-discrete measure and a discrete measure}
\author{Ying LU\inst{1} \and Liming CHEN\inst{1} \and Alexandre SAIDI\inst{1} \and Xianfeng GU\inst{2}}

\institute{Ecole Centrale de Lyon, France
\and 
Stony Brook University, USA}

\date{}

\begin{document}
\maketitle

\begin{abstract}
Correctly estimating the discrepancy between two data distributions has always been an important task in Machine Learning. Recently, Cuturi proposed the Sinkhorn distance \cite{cuturi2013sinkhorn} which makes use of an approximate Optimal Transport cost between two distributions as a distance to describe distribution discrepancy. Although it has been successfully adopted in various machine learning applications (\emph{e.g.} in Natural Language Processing and Computer Vision) since then, the Sinkhorn distance also suffers from two unnegligible limitations. The first one is that the Sinkhorn distance only gives an approximation of the real Wasserstein distance, the second one is the `divide by zero' problem which often occurs during matrix scaling when setting the entropy regularization coefficient to a small value. In this paper, we introduce a new Brenier approach for calculating a more accurate Wasserstein distance between two discrete distributions, this approach successfully avoids the two limitations shown above for Sinkhorn distance and gives an alternative way for estimating distribution discrepancy.
\end{abstract}

\section{Introduction}

In Machine Learning and Pattern Recognition, the data we always work with are samples. For example in image classification, a sample is an image. Given a fixed type of representation for images, we can consider a space of images in which each dimension represents a feature (\emph{e.g.} a pixel) of image. In this way, we can consider a certain set of images (\emph{e.g.} a set of images correspond to a semantic concept like `cat') as a distribution (or a measure) over the space of images. 

In this situation, it is not easy to estimate this kind of distributions as a continuous probability distribution. Firstly because we usually only have access to a finite number of training samples for a certain set (or a category), compared to the large number of dimensions in the image space, these training samples are always not enough for estimating a continuous probability function \footnote{\url{https://en.wikipedia.org/wiki/Curse_of_dimensionality}}. Secondly, when the categories are defined as semantic concepts which are highly abstract, we cannot ensure that they are continuous distributions by nature. For example, if we do interpolation in the space of images between two images of cat, we are not guaranteed to get a new image which can be seen as an image of cat in human eyes (when the two input images are very different, usually what we can get is a image of `noise'). 

Therefore, in Machine Learning society, the commonly adopted methods for measuring distance between two distributions are usually sample based methods (\emph{e.g.} two-sample test methods as MMD \cite{gretton2012kernel}) or consider distributions as discrete measures (\emph{e.g.} sinkhorn distance \cite{cuturi2013sinkhorn}). 

In this paper we discuss the application of the Brenier approach \cite{gu2013variational} for calculating the optimal transportation between a quasi-discrete measure and a discrete measures (a discrete measure can be represented with a finite number of points, each point corresponds to a Dirac measure). The approach introduced in \cite{gu2013variational} assumes that the target measure is discrete while the source measure is continuous. To estimate the Brenier function and the gradient of objective energy, we need to estimate the Graph of Brenier potential and calculate integration in the source distribution. This could be done for one dimensional or two dimensional spaces in a reasonable time of calculation, while becomes hard for spaces with more than two dimensions. 

Therefore, by considering the source measure as a quasi-discrete distribution, we wish to find a way to solve the discrete problems in machine learning and at the same time to avoid the problem of calculation cost for higher dimensional spaces. 
However the side effect is that when the quasi-discrete measure gets close to a discrete measure, 
we are not always guaranteed to converge to a solution that preserves the measures. 

In the following part of this paper, we firstly introduce the Brenier approach for optimal transportation between a quasi-discrete measure and a discrete measure, and the Gradient Descent algorithm for solving this problem. We then compare the Brenier approach with sinkhorn approach \cite{cuturi2013sinkhorn}, discuss their advantages and disadvantages. We also show a possible application of Brenier approach for clustering.

\section{Brenier approach for OMT between a quasi-discrete and a discrete measure}
\label{sec:brenier_def}

Assume $\mu$ and $\nu$ two discrete measures represented by two sample sets: $\{\mathbf{x}^s_1, \ldots, \mathbf{x}^s_{n_s}\}$ (source sample set) and $\{\mathbf{x}^t_1, \ldots, \mathbf{x}^t_{n_t}\}$ (target sample set), in the $n$-dimensional Euclidean space $\mathbb{R}^n$: 

\begin{equation}
\label{eq:dirac_mu_nu}
\mu = \sum^{n_s}_{i=1} p^s_i \delta (\mathbf{x}-\mathbf{x}^s_i),\ \ \nu = \sum^{n_t}_{i=1} p^t_i \delta (\mathbf{x}-\mathbf{x}^t_i)
\end{equation}

 where $\delta (\mathbf{x}-\mathbf{x}_i)$ is the Dirac function at location $\mathbf{x}_i$, $p^s_i$ and $p^t_i$ are probability masses associated to the $i$-th sample in source set and target set respectively, and $\sum^{n_s}_{i=1} p^s_i = \sum^{n_t}_{i=1} p^t_i$. 
 
 Given a cost function $c:\ \mathbb{R}^n \times \mathbb{R}^n \to \mathbb{R}$, the Monge's optimal transport problem is to find the unique measure preserving map $T:\ \mathbb{R}^n \to \mathbb{R}^n$ (from $\mu$ to $\nu$) that minimizes the total transportation cost:
 
 \begin{equation}
 \label{eq:omt}
 \mathcal{C}(T):= \int_{\mathbb{R}^n} c(\mathbf{x}, T(\mathbf{x})) \mathrm{d} \mu(\mathbf{x})
 \end{equation}
 
 The theorem of Brenier and the variational approach in \cite{gu2013variational} assumes that the source measure is absolutely continuous (with respect to Lebesgue measure) and the support of source measure is a convex set in $\mathbb{R}^n$, while this is not true for $\mu$. Therefore we introduce a piecewise uniform measure $\mu'$ with a compact support set $\Omega$, which could be seen as an approximation to $\mu$. The probability density function of $\mu'$ is defined as:
 
 \begin{equation}
 \label{eq:pdf_mu_uniform}
 f_{\mu'}(\mathbf{x}) = \begin{cases}
 \frac{p^s_i-p_0/n_s}{\varepsilon^n} & \text{for } \mathbf{x} \in [(\mathbf{x}^s_i)_1-\frac{\varepsilon}{2}, (\mathbf{x}^s_i)_1+\frac{\varepsilon}{2}] \times \ldots \times [(\mathbf{x}^s_i)_n-\frac{\varepsilon}{2}, (\mathbf{x}^s_i)_n+\frac{\varepsilon}{2}],\forall \ i \in \{1, \ldots, n_s\} \\
 \frac{p_0}{vol(\Omega)-n_s \varepsilon^n} & \text{for } \mathbf{x} \in \Omega \text{ elsewhere.}
 \end{cases}
 \end{equation}
 
 where $\varepsilon$ and $p_0$ are very small values. The probability density is uniformly distributed in a small hypercube of volume $\varepsilon^n$ around each source sample, and the total probability mass in a small hypercube is defined as the probability mass associated to the center sample minus a small value $p_0/n_s$. Apart from the small hypercubes around source samples, probability is uniformly distributed in $\Omega$, and the total mass in the rest volume is $p_0$. $\Omega$ could be defined as the smallest hypercube area which contains all source samples.
 
 With this measure $\mu'$, we can now apply the theorem of Brenier (Theorem 9.4 in \cite{villani2008optimal}):  
 \emph{Let $c(\mathbf{x},\mathbf{x}') = \vert \mathbf{x} - \mathbf{x}' \vert^2$ in $\mathbb{R}^n$, There exists a convex function $f: \mathbb{R}^n \to \mathbb{R}$, its gradient map $\nabla f$ gives the solution to the Monge's problem (from $\mu'$ to $\nu$), and this map is unique.} This convex function is called the Brenier potential, and it should be a solution to the Monge-Ampère equation. In \cite{gu2013variational} the authors give a variational approach to solve this equation with the equivalent Alexandrov Theorem. We now introduce this approach:
 
 Define a vector $\mathbf{h} = (h_1, \ldots, h_{n_t}) \in \mathbb{R}^{n_t}$. For each target sample $\mathbf{x}^t_i$, we define a hyperplane $\pi_i: \langle \mathbf{x}, \mathbf{x}^t_i\rangle + h_i = 0$ in $\mathbb{R}^n$, 
 the upper envelope of all the hyperplanes forms a piecewise linear convex function:
 
 \begin{equation}
 \label{eq:graph}
 u_{\mathbf{h}}(\mathbf{x}) = \max^{n_t}_{i=1}\{ \langle \mathbf{x}, \mathbf{x}^t_i \rangle + h_i \}
 \end{equation}
 
 Denote its graph by $G(\mathbf{h})$, which is an infinite convex polyhedron with supporting planes $\pi_i(\mathbf{h})$. The projection of $G(\mathbf{h})$ induces a polygonal partition of $\Omega$, where each cell $W_i(\mathbf{h})$ is the projection of a facet of $G(\mathbf{h})$ onto $\Omega$. The area of each cell is defined as:
 
 \begin{equation}
 \label{eq:area_cell}
 w_i(\mathbf{h}) = \int_{W_i(\mathbf{h})\cap \Omega} f_{\mu'}(\mathbf{x})\textrm{d}
 \mathbf{x}
 \end{equation}
 
 The convex function $u_{\mathbf{h}}$ on each cell $W_i(\mathbf{h})$ is a linear function $\pi_i(\mathbf{h})$, therefore, the gradient map:
 
 \begin{equation}
 \label{eq:grad_map}
 grad\ u_\mathbf{h}:\ \ W_i \to \mathbf{x}^t_i,\ \ \forall i \in \{1, \ldots, n_t\}
 \end{equation}
 
 maps each area $W_i(\mathbf{h})$ to a single point $\mathbf{x}^t_i$.
 The problem is to find a vector $\mathbf{h}$ such that the polygonal partition $\{W_i\}^{n_t}_{i=1}$ of the source support $\Omega$ induced by the projection of $G(\mathbf{h})$ on $\mathbb{R}^n$ is measure preserving.
 In \cite{gu2013variational} the authors prove that the solutions of this problem are the critical points of the following energy function:
 
 \begin{equation}
 \label{eq:energy}
 E(\mathbf{h}) = \int^{\mathbf{h}} \sum^{n_t}_{i=1} w_i(\mathbf{h}) \mathrm{d}h_i - \sum^{n_t}_{i=1} p^t_i h_i
 \end{equation}
 
 The first part of this energy function is the volume of the area bounded by the graph $G(\mathbf{h})$, the horizontal plane $\{y=0\}$, and the cylinder consisting of vertical lines through $\partial\Omega$.
 
 
 The gradient of this energy function with respect to $\mathbf{h}$ is:
 
 \begin{equation}
 \label{eq:gradient_E}
 \frac{\partial E(\mathbf{h})}{\partial h_i} = w_i(\mathbf{h}) - p^t_i, \ \ \ \forall i \in \{1, \ldots, n_t\}
 \end{equation}
 
 In \cite{gu2013variational} the authors prove that \emph{when $\Omega$ is convex, the admissible space $H_0$ for $\mathbf{h}$ is convex, so is the energy in Eq. \eqref{eq:energy}. Moreover, the unique global minimum $\mathbf{h}_0$ is an interior point of $H_0$. And the gradient map Eq. \eqref{eq:grad_map} induced by the minimum $\mathbf{h}_0$ is the unique optimal mass transport map, which minimizes the total transportation cost Eq. \eqref{eq:omt} with $c(\mathbf{x},\mathbf{x}') = \vert \mathbf{x} - \mathbf{x}' \vert^2$.}
 
 Since Eq. \eqref{eq:energy} is convex, we can therefore use a gradient descent approach to solve this problem. 
 
 Furthermore, since for source measure $\mu'$ the probability masses are concentrated in small areas around the source samples, we wish to simplify the calculation by estimating $G(\mathbf{h})$ with only the source samples instead of all possible points in $\Omega$. Therefore, we can define the approximation of $u_{\mathbf{h}}$ in Eq. \eqref{eq:graph} as follows:
 
 \begin{equation}
 \label{eq:graph_approxi}
 \hat{u}_{\mathbf{h}}(\mathbf{x}^s_i) = \max^{n_t}_{j=1} \{\langle \mathbf{x}^s_i, \mathbf{x}^t_j \rangle + h_j\},\ \ \forall i \in \{1, \ldots, n_s\}
 \end{equation}
 
 A possible problem of Eq. \eqref{eq:graph_approxi} is that there might be some source sample, for example $\mathbf{x}^s_k$, for which the corresponding point $[\mathbf{x}^s_k,\hat{u}_{\mathbf{h}}(\mathbf{x}^s_k)]$ is situated on the intersection of multiple hyperplanes. In other words, the size of the following set (which is a subset of target sample set) is larger than 1.
 
 \begin{equation}
 \label{eq:mapping_func}
 t(\mathbf{x}^s_k) = \{\mathbf{x}^t_j\ \vert\ \langle \mathbf{x}^s_k, \mathbf{x}^t_j \rangle + h_j = \hat{u}_{\mathbf{h}}(\mathbf{x}^s_k)\}
 \end{equation}
  
  This will make it difficult to calculate cell areas. To solve this problem, currently we use a brute force method by simply uniformly distribute the mass correspond to this source sample $\mathbf{x}^s_k$ onto all cell areas correspond to the target samples in $t(\mathbf{x}^s_k)$. 
 
 With the brute force method described above, we can therefore approximately calculate the area of each projected cell $W_j(\mathbf{h})$ as follows instead of using Eq. \eqref{eq:area_cell}:
 
 \begin{equation}
 \label{eq:area_cell_approxi}
 \hat{w}_j = \sum_{\mathbf{x}^s_i \in W_j(\mathbf{h})} \frac{1}{\vert t(\mathbf{x}^s_i) \vert} p^s_i
 \end{equation}
 
 where $\vert t(\mathbf{x}^s_i) \vert$ is the size of set $t(\mathbf{x}^s_i)$. The transportation map induced by this approximation of Brenier potential can be expressed as a matrix $\mathbf{T}$ of size $n_s \times n_t$ with each element defined as follows:
 
 \begin{equation}
 \label{eq:transp}
 T_{ij} = \begin{cases}
 \frac{1}{\vert t(\mathbf{x}^s_i) \vert} p^s_i & \textrm{ if } \mathbf{x}^s_i \in W_j(\mathbf{h}) \\
 0 & \textrm{ otherwise}
 \end{cases}
 \end{equation}
 
 The energy function can also be approximately calculated instead of Eq. \eqref{eq:energy}:
 
\begin{equation}
\label{eq:energy_approxi}
 \hat{E}(\mathbf{h}) = \sum^{n_t}_{j=1} \sum_{\mathbf{x}^s_i \in W_j(\mathbf{h})} \frac{1}{\vert t(\mathbf{x}^s_i) \vert} p^s_i \hat{u}_{\mathbf{h}}(\mathbf{x}^s_i) - \sum^{n_t}_{j=1} p^t_j h_j
\end{equation}

The gradient descent algorithm to solve this problem can then be defined as follows:
 
 \begin{algorithm}
 \renewcommand{\algorithmicrequire}{\textbf{Input:}}
 \renewcommand\algorithmicensure {\textbf{Output:} }
\caption{Gradient Descent Algorithm for approximately solving OMT}
\label{algo:gd}
\begin{algorithmic}[1]
\REQUIRE 
Source sample set: $\{\mathbf{x}^s_1, \ldots, \mathbf{x}^s_{n_s}\}$; Target sample set: $\{\mathbf{x}^t_1, \ldots, \mathbf{x}^t_{n_t}\}$; Number of steps $N$; Step size $\lambda$.
\STATE 
Initialize: Vector $\mathbf{p}^s$ of size $n_s$, where $p^s_i = \frac{1}{n_s}$; Vector $\mathbf{p}^t$ of size $n_t$, where $p^t_i = \frac{1}{n_t}$; \\
\ \ \ \ \ \ \ \ \ \ \ \ \ Vector $\mathbf{h}$ of size $n_t$, where $h_i = 0$; \\
\ \ \ \ \ \ \ \ \ \ \ \ \ Inner product matrix $\mathbf{M}$ of size $n_s \times n_t$, where $Mij = \langle \mathbf{x}^s_i, \mathbf{x}^t_j \rangle$; \\
\ \ \ \ \ \ \ \ \ \ \ \ \ Gradient vector $\mathbf{g}$ of size $n_t$ where $g_i=0$;\\
\ \ \ \ \ \ \ \ \ \ \ \ \ Counter $n_{step}=0$
\WHILE{($\mathbf{h}$ not converged) and ($n_{step} < N$)}
\STATE
Gradient Descent: $\mathbf{h} = \mathbf{h} - \lambda \mathbf{g}$.
\STATE
(This step could be done with a series of matrix calculation with $\mathbf{M}$.) \\
Update $\hat{u}_{\mathbf{h}}(\mathbf{x}^s_i)$ with Eq. \eqref{eq:graph_approxi} for all source samples. \\
Update: Cell areas $\{\hat{w}_j\}^{n_t}_{j=1}$ with Eq. \eqref{eq:area_cell_approxi}. 
\STATE
Update: gradient vector $\mathbf{g}$ where $g_j = \hat{w}_j - p^t_j$.
\STATE
Update counter: $n_{step} = n_{step}+1$.
\ENDWHILE
\STATE
Calculate transportation map $\mathbf{T}$ with Eq. \eqref{eq:transp}.
\ENSURE 
Transportation map $\mathbf{T}$.
\end{algorithmic}
 \end{algorithm}
 
 With the resulting transportation plan $\mathbf{T}$, we can further calculate the transportation cost (\emph{i.e.} the Wasserstein distance) as follows:
 
 \begin{equation}
 \label{eq:brenier_wdist}
 W(\mu', \nu) = \sum^{n_s}_{i=1} \sum^{n_t}_{j=1} T_{ij} \vert \mathbf{x}^s_i - \mathbf{x}^t_j \vert^2
 \end{equation}
 
\section{Brenier v.s. Sinkhorn}
\label{sec:brenier_vs_sinkhorn}

In the previous section, we have introduced an approximate approach to find the Brenier potential in order to solve the Optimal Transportation problem. This approximate Brenier approach solves the same kind of problems as considered in \cite{cuturi2013sinkhorn}. Therefore in this section we compare the proposed approximate Brenier approach with the Sinkhorn approach. The differences are listed as follows:

\begin{enumerate}
\item {The two methods approach the optimal solution in different ways. 

In each step of the approximate Brenier method, the condition 
$\mathbf{T} \mathbf{1}_{n_t} = \mu$ always holds\footnote{Here $\mathbf{1}_d$ is a $d$-dimensional column vector of ones.}, while $\hat{\nu} = \mathbf{T}^{\top} \mathbf{1}_{n_s}$ is approaching the real $\nu$ when the number of iteration grows.

On the other hand, the Sinkhorn method is a matrix balancing method, the two conditions $\mathbf{T} \mathbf{1}_{n_t} = \mu$ and $\mathbf{T}^{\top} \mathbf{1}_{n_s} = \nu$ hold alternatively during iterations (\emph{i.e.} the first condition holds after updating the vector $v$, and the second condition holds after updating the vector $u$), and they tend to both hold when the algorithm converges.}

\item {In each iteration step of the approximate Brenier method, the current transportation map is always an optimal transportation from $\mu'$ to $\hat{\nu} = \mathbf{T}^{\top}\mathbf{1}_{n_s}$.}


\item {The Sinkhorn method solves a entropy regularized version of the OT problem. Therefore we can only get the optimal solution of the original OT problem when the coefficient $\lambda$ of the regularization term tends to zero. However this is hard to achieve because when $\lambda$ tends to zeros, the matrix balancing becomes instable and we are more easily to face a zero denominator error. In other words, the entropy regularized OT problem demands the transportation plan $\mathbf{T}$ to be not sparse (with no zeros in it), since to use matrix balancing we need the matrix to be with all positive entries. 

Fortunately, we don't have this problem with the approximate Brenier method.}

\item {The transportation map learned by Brenier method tend to transport each source sample as a whole to some target sample (the situation mentioned in section \ref{sec:brenier_def} where a source sample is situated on the intersection of two cells is actually rare in practice.) 

On the other hand, the transportation map learned by Sinkhorn method tend to split a source sample into parts and transport it to a group of target samples.}

\item {One thing in common about the two methods is that they both can handle abstract distributions, \emph{i.e.} we don't need to know the number of dimensions of the sample space.

For Sinkhorn all we need to prepare are two vectors of sample weights and a distance matrix (the cost matrix).

For approximate Brenier all we need to prepare are two vectors of sample weights and a inner-product matrix (the matrix $\mathbf{M}$ defined in Algorithm \ref{algo:gd}, step 1).

This characteristic makes the two methods flexible for different kinds of applications.}

\item {The Sinkhorn method is much faster than the Brenier method.} 

\item {The approximate Brenier method is very hard to converge (sometimes impossible to converge) unless the number of source samples is much larger than the number of target samples.

No such problem for Sinkhorn method.}


\item {The Sinkhorn method has no constraints on the choice of cost metric, while currently the Brenier method only works for quadratic Euclidean distances.} 

\item {A simple experimentation to compare the two methods: 
the source sample set has 150 samples from a Gaussian mixture distribution, the target sample set has only 2 samples. We set the Sinkhorn regularization coefficient $\lambda=0.05$, and we use quadratic Euclidean distance as cost metric for both methods, then the resulting Sinkhorn distance is 2053.47, and the calculation time is 0.00335 seconds; the Brenier distance is 2015.08, and the calculation time is 5.927 seconds. We can see that the Brenier distance is smaller than the Sinkhorn distance, meaning that the transportation map learned by Brenier method is better than that learned by Sinkhorn method. We also performed a simple linear programming method to solve this problem, the resulting distance is 2015.08 (the same as that with Brenier method), and the calculation time is 0.094 seconds.

We perform a even smaller experiment to show the maps learned by different methods: consider a source set of 10 samples from a Gaussian mixture distribution, and a target set with only 2 samples. Use quadratic Euclidean distance as cost metric. Set $\lambda=0.05$ for Sinkhorn method. The results are shown in the following, where $W$ means the resulting Wasserstein distance, $\mathbf{T}$ is the transportation map, `B' represents Brenier, `S' represents Sinkhorn and `LP' represents linear programming. We can see from the results that only the Brenier method learns the simple and elegant optimal transportation map.

$$
W_B = 2065.694\ \ \ \ \ \ \ \ W_S = 2116.271\ \ \ \ \ \ \ \ W_{LP} = 2065.694
$$
$$
\mathbf{T}_{B} = \begin{bmatrix}
0.1 & 0. \\
0.1 & 0. \\
0.1 & 0. \\
0.1 & 0. \\
0.1 & 0. \\
0.  & 0.1\\
0.  & 0.1\\
0.  & 0.1\\
0.  & 0.1\\
0.  & 0.1\\
\end{bmatrix}\ \ 
\mathbf{T}_{S} = \begin{bmatrix}
0.66414413 & 0.33585587\\
0.66696907 & 0.33303093\\
0.69859367 & 0.30140633\\
0.7877272  & 0.2122728 \\
0.72616414 & 0.27383586\\
0.30988759 & 0.69011241\\
0.33522328 & 0.66477672\\
0.22635254 & 0.77364746\\
0.27228676 & 0.72771324\\
0.31265155 & 0.68734845\\
\end{bmatrix}$$
$$
\mathbf{T}_{LP} = \begin{bmatrix}
9.99999865\times 10^{-2} &  1.35073998\times 10^{-8}\\
9.99999868\times 10^{-2} &  1.32129803\times 10^{-8}\\
9.99999926\times 10^{-2} &  7.35148907\times 10^{-9}\\
9.99999947\times 10^{-2} &  5.25121156\times 10^{-9}\\
9.99999984\times 10^{-2} &  1.58285275\times 10^{-9}\\
1.00561170\times 10^{-8} &  9.99999899\times 10^{-2}\\
1.25810968\times 10^{-8} &  9.99999874\times 10^{-2}\\
4.32183817\times 10^{-9} &  9.99999957\times 10^{-2}\\
3.44685417\times 10^{-9} &  9.99999966\times 10^{-2}\\
1.05000275\times 10^{-8} &  9.99999895\times 10^{-2}\\
\end{bmatrix}
$$}
\end{enumerate}

\section{Brenier approach for clustering}
\label{sec:b_clustering}

Inspired by the comparisons shown in the previous section, I think a good way to apply this approximate Brenier approach is to apply it for clustering (because in this case the source samples will be much more than the target samples and the Brenier method is much easier to converge). 

For a clustering task, we are given an unlabeled sample set and we are demanded to divide this sample set into clusters, where the samples in a same cluster should be close to each other while samples from different clusters should be far from each other. We can assume the given sample set as the source set $\{\mathbf{x}^s_i\}$, and assume the set of cluster centers, which we need to learn, as the target set $\{\mathbf{x}^t_j\}$. In this case, both the transportation map and the target distribution are unknown variables. We can therefore use an iterative approach (like EM algorithm) where we firstly initialize the target samples $\{\mathbf{x}^t_j\}$ randomly, then we learn the intercepts $\mathbf{h}$ and the target samples $\{\mathbf{x}^t_j\}$ alternatively. 

The objective for learning cluster centers $\{\mathbf{x}^t_j\}$ is to minimize the Wasserstein distance between the given samples and the cluster centers, \emph{i.e.} to minimize Eq. \eqref{eq:brenier_wdist}. Since $\mathbf{T}$ is a matrix with positive elements, the Eq. \eqref{eq:brenier_wdist} is convex with respect to input set $\{\mathbf{x}^t_j\}$. And its gradients with respect to $\{\mathbf{x}^t_j\}$ are:

\begin{equation}
\label{eq:b_wdist_gradients}
\frac{\partial W}{\partial \mathbf{x}^t_j} = \sum^{n_s}_{i=1} 2 T_{ij}(\mathbf{x}^t_j - \mathbf{x}^s_i)
\end{equation}

We can therefore use a simple gradient descent algorithm to learn $\{\mathbf{x}^t_j\}$.

\begin{figure}[!htb]
\centering
\includegraphics[width=0.6\linewidth]{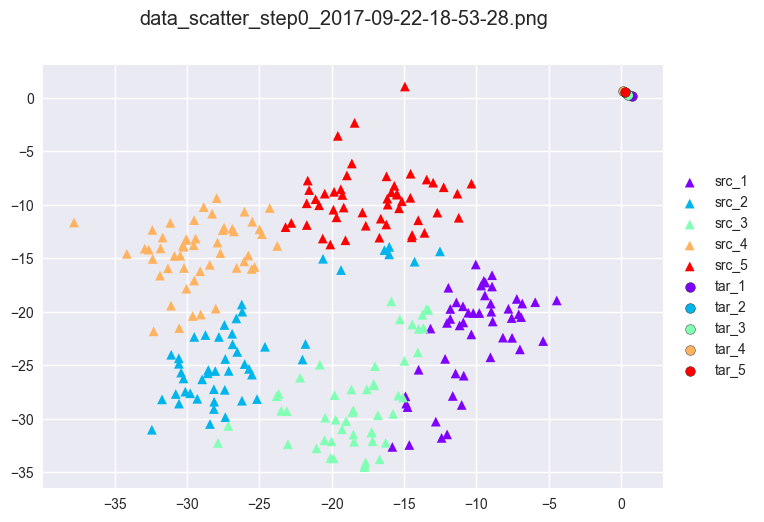}
\caption{Approximate Brenier method for clustering: initialization}
  \label{fig:b_clustering_init}
\end{figure}
\begin{figure}[!htb]
  \centering
  \includegraphics[width=0.46\linewidth]{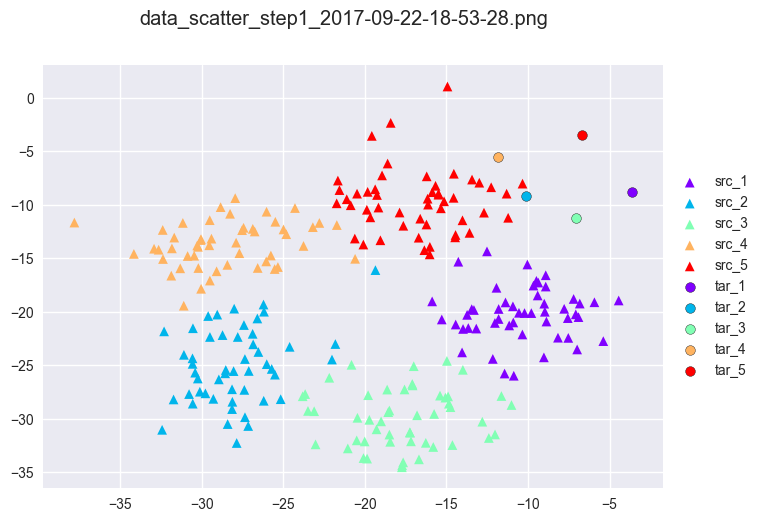}
  \includegraphics[width=0.46\linewidth]{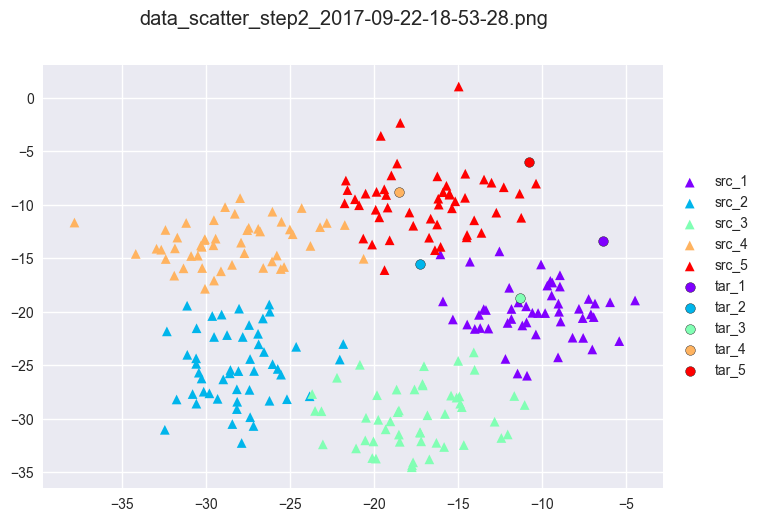}
  \includegraphics[width=0.46\linewidth]{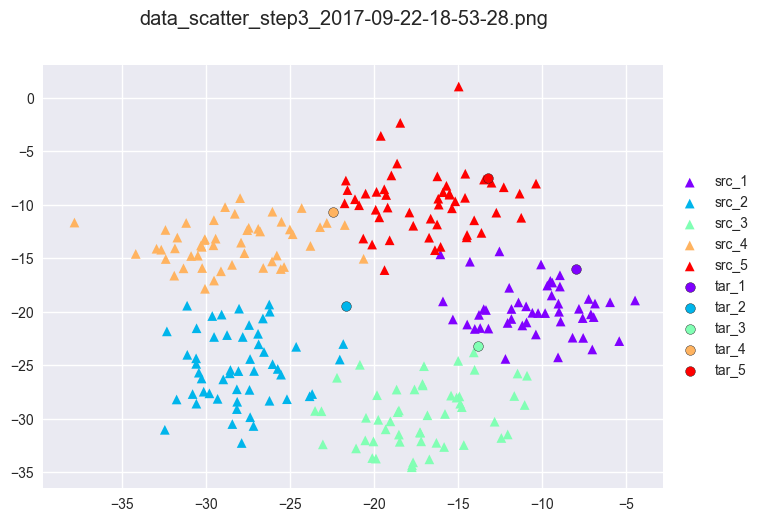}
  \includegraphics[width=0.46\linewidth]{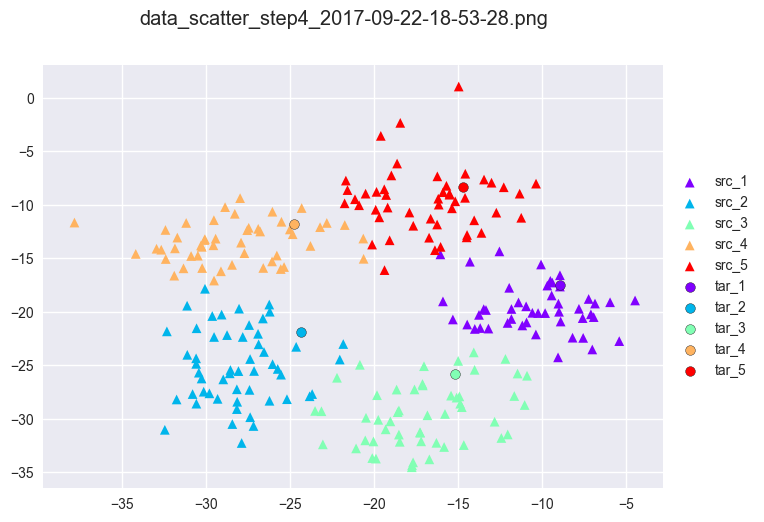}
  \includegraphics[width=0.46\linewidth]{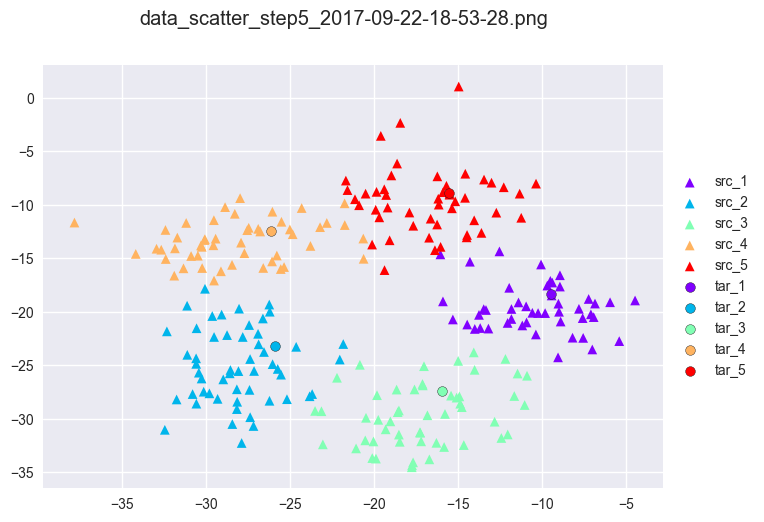}
  \includegraphics[width=0.46\linewidth]{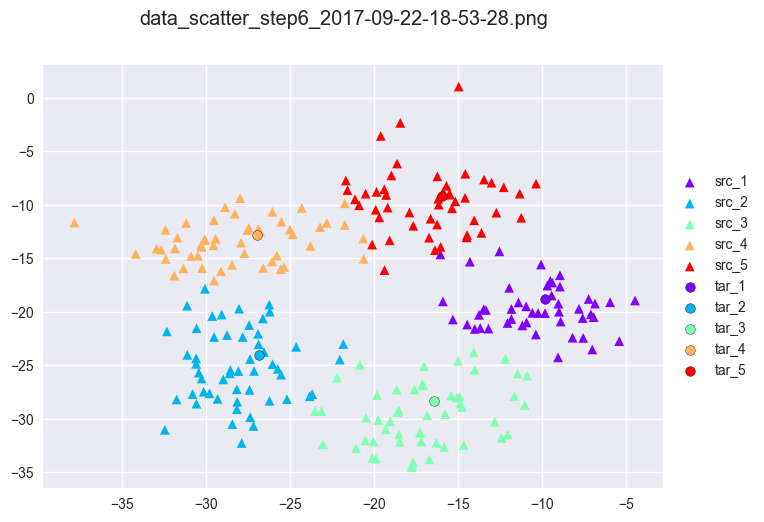}
  \includegraphics[width=0.46\linewidth]{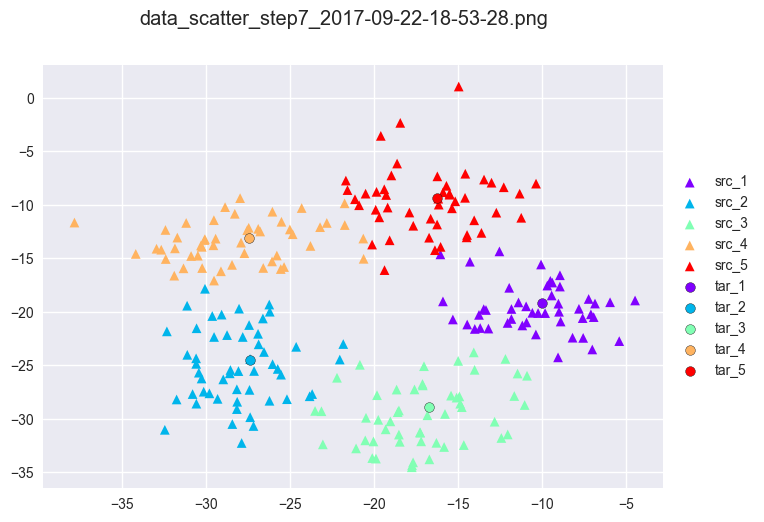}
  \includegraphics[width=0.46\linewidth]{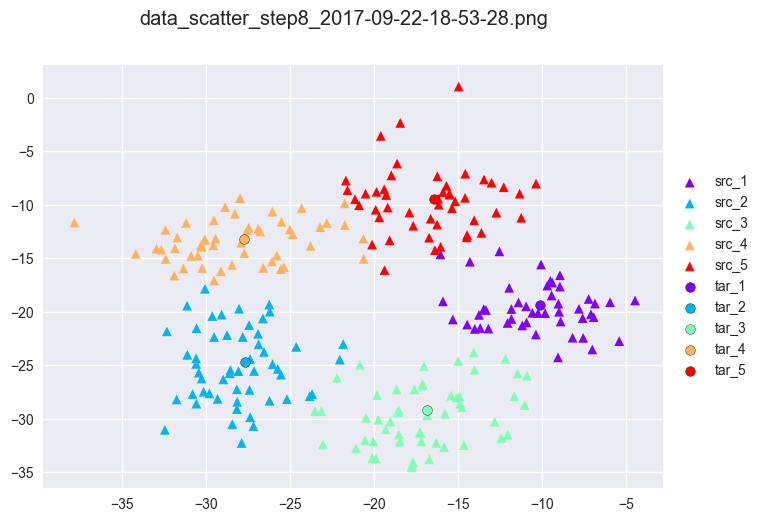}
  \includegraphics[width=0.46\linewidth]{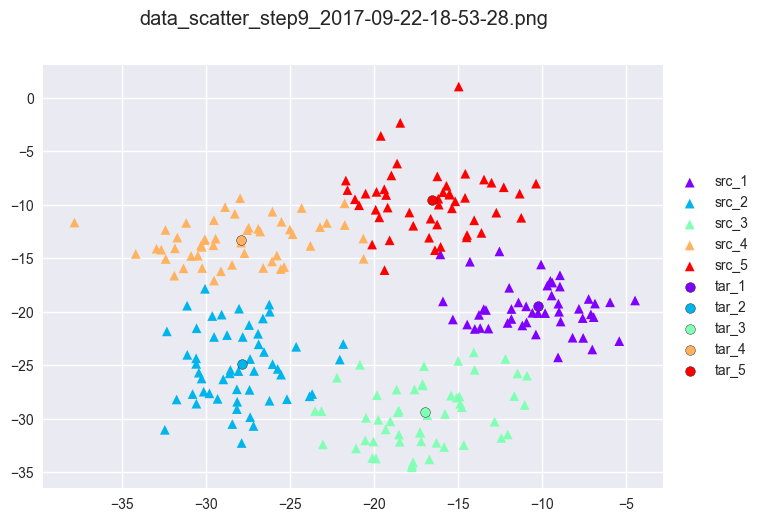}
  \includegraphics[width=0.46\linewidth]{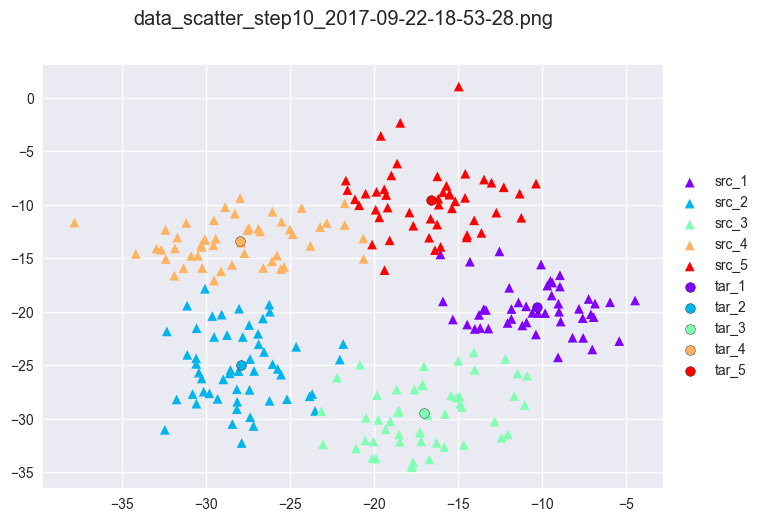}
  \caption{Approximate Brenier method for clustering: step 1 to 10}
  \label{fig:b_clustering_steps}
\end{figure}

We show a simple experiment in the following: given a set of 250 samples (2D points) from a Gaussian mixture distribution (5 Gaussians), we firstly initialize 5 cluster centers randomly, and assume the distribution mass associated to each cluster center is $0.2$, then we perform 10 steps of gradient descent for updating cluster centers, where in each step we perform the approximate Brenier method to learn the current optimal transportation map. In each step, the samples which are transported to a same cluster center are considered as a cluster. In figure \ref{fig:b_clustering_init} we show the initialized clusters (src\_1 - src\_5) and cluster centers (tar\_1 - tar\_5). In figure \ref{fig:b_clustering_steps} we illustrate the resulting clusters and corresponding centers in each step. We can see that in only 10 steps the learned cluster centers are nicely located in the center of each Gaussian distribution. 

\section{Conclusion}
\label{sec:discuss}

In this paper we show a simple application of the Brenier approach for approximately solving discrete learning problems. This method is not as time consuming as continues Brenier approach, but is still slower than other methods like Sinkhorn or even linear programming. Although currently it is not perfect, we can still find some shining points in it (as we have discussed in section \ref{sec:brenier_vs_sinkhorn}). Therefore we hope to further improve this method and find more applications in machine learning and computer vision.

\bibliographystyle{ieeetr}
\bibliography{sample}

\end{document}